\documentclass[journal]{IEEEtran}

\usepackage{graphicx}
\usepackage{multirow}
\usepackage{amsmath}
\usepackage{algorithmic}
\usepackage[ruled]{algorithm2e}
\usepackage{bm}
\usepackage{amsfonts}
\usepackage{subfigure}
\usepackage{booktabs}

\title{Offline Signature Verification Based on Feature Disentangling Aided Variational Autoencoder}

\author{
	\begin{minipage}[t]{0.33\textwidth}
		\centering
		\IEEEauthorblockN{Hansong Zhang} \\
		\IEEEauthorblockA{
			\textit{University of California,\\ San Diego}\\ La Jolla, CA, US\\
			haz064@ucsd.edu
		}
	\end{minipage}%
	\hfill
	\begin{minipage}[t]{0.33\textwidth}
	\centering
	
	\IEEEauthorblockN{Jiangjian Guo} \\
	\IEEEauthorblockA{
		\textit{University of California,\\ San Diego}\\ La Jolla, CA, US\\
		j9guo@ucsd.edu
	}
	\end{minipage}
	\begin{minipage}[t]{0.33\textwidth}
		\centering
		\IEEEauthorblockN{Kun Li} \\
		\IEEEauthorblockA{
		\textit{University of Illinois Urbana-Champaign}\\ Champaign, IL, US\\
		kunli3@illinois.edu
	}
	\end{minipage}

	\vspace{0.5cm}
	
	\begin{minipage}[t]{0.5\textwidth}
		\centering
		
		\IEEEauthorblockN{Yang Zhang} \\
		\IEEEauthorblockA{
			\textit{Boston University}\\ Boston, MA, US\\
			ycheung@bu.edu
		}
	\end{minipage}%
	\hfill
	\begin{minipage}[t]{0.5\textwidth}
		\centering
		\IEEEauthorblockN{Yimei Zhao} \\
		\IEEEauthorblockA{
			\textit{University of California,\\ San Diego}\\ La Jolla, CA, US\\
			yiz049@ucsd.edu
	}
	\end{minipage}
	\hfill
}

\begin{document}
	
\maketitle

\begin{abstract}
Offline handwritten signature verification systems are used to verify the identity of individuals, through recognizing their handwritten signature image as genuine signatures or forgeries. The main tasks of signature verification systems include extracting features from signature images and training a classifier for classification. The challenges of these tasks are twofold. First, genuine signatures and skilled forgeries are highly similar in their appearances, resulting in a small inter-class distance. Second, the instances of skilled forgeries are often unavailable, when signature verification models are being trained. To tackle these problems, this paper proposes a new signature verification method. It is the first model that employs a variational autoencoder (VAE) to extract features directly from signature images. To make the features more discriminative, it improves the traditional VAEs by introducing a new loss function for feature disentangling. In addition, it relies on SVM (Support Vector Machine) for classification according to the extracted features. Extensive experiments are conducted on two public datasets: MCYT-75 and GPDS-synthetic where the proposed method significantly outperformed $13$ representative offline signature verification methods. The achieved improvement in distinctive datasets indicates the robustness and great potential of the developed system in real application.
\end{abstract}



\IEEEpeerreviewmaketitle

\section{Introduction}
Offline handwritten signature verification aims at recognizing a signature image as genuine or forgery~\cite{cnn2}. Compared with other biometric traits (e.g., iris and fingerprint), handwritten signatures are non-invasive and more convenient to collect in practice. Therefore, offline handwritten signature verification systems have been widely used in a wide range of application scenarios comparing to its online counterpart \cite{2c2s}. The goal of handwritten signature verification systems is to accept genuine signatures and reject skilled forgeries. In practice, however, these two kinds of signature images are highly similar in their appearances. For illustration, Fig. 1 provides the instances of genuine signatures and skilled forgeries of three users. Obviously, it is usually difficult for a nonspecialist to distinguish them. 


 \begin{figure}
	\centering
	\scalebox{0.20}[0.20]{\includegraphics {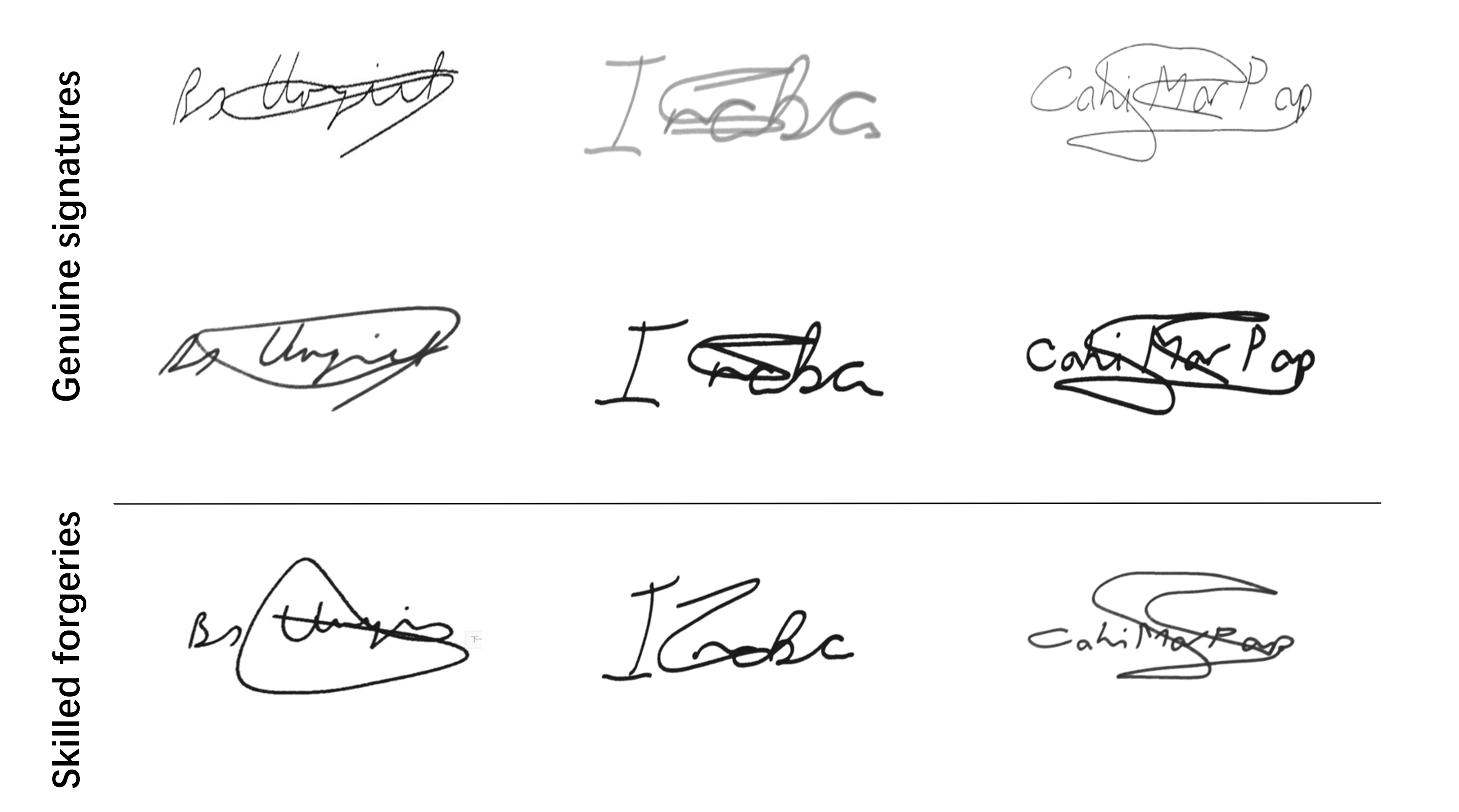}}
	\caption{Genuine signatures vs. Skilled forgeries. Each column has two genuine signatures and one skilled forgery for one user.}
\end{figure}

The existing offline handwritten signification verification systems usually rely on the machine learning based methods. They train a classifier for binary classification (i.e., genuine or forgery) according to the features extracted from signature images. There exist two main challenges in building signature verification systems. First, as shown in fig. 1, the inter-class variance is small, but the intra-class variance is relatively large \cite{blank1}. This requires the features extracted from signature images should be sufficiently discriminative. Second, skilled forgeries are often unavailable during model training. Hence, the classifier has limited knowledge about skilled forgeries. 

In this paper, we propose a new signature verification method that successfully overcomes these two challenges. Our method employs an improved \textit{variational autoencoder} (VAE) to learn the features directly from signature images and uses SVM for classification. The intuition behind our VAE-based feature extraction is described as follows. First, if our VAE model is trained to learn underlying distribution of signatures and accurately reconstruct signatures, it will be able to endure large intra-class variance. Second, we improve the traditional VAE based on the idea of \textit{feature disentangling}, in order to enhance its capability of discriminating genuine signatures from skilled forgeries. In this way, we can alleviate the problem of small inter-class variance. Last, our method only needs the instances of genuine signatures and random forgeries for model training\footnote{Without the knowledge about the victim user, one uses the signatures of the other users as the random forgeries.}, without requiring the instances of skilled forgeries. 

In summary, the main contributions of this work include:
\begin{itemize}
	\item To our knowledge, this is the first work that uses a VAE to extract features for offline handwritten signature verification.
	\item We improve the existing VAEs by introducing a new loss function to disentangle the features of different classes, which makes the extracted features more discriminative.  
	\item Our method achieves promising results on two public datasets, significantly outperforming $13$ representative signature verification methods. 
\end{itemize}

\section{Related Work}
In general, the performance of signature verification relies heavily on the features extracted from signature images\cite{Black}, which can be roughly divided into two categories. 
\subsubsection{Handcrafted Features}The traditional methods often use the handcrafted features. For instance, in \cite{global}, global features, such as heights, widths, area and aspect ratio are utilized to extract discernible patterns.  In \cite{ga}, the proposed method considers both global features (e.g., the area of signature) and local features (e.g., slope), and selects the best ones with a genetic algorithm. BoVM-VLAD-KAZE \cite{BVK} uses KAZE features detected in strokes and background. The method proposed in \cite{GLCM} uses Gray Level Co-occurrences Matrix and fused by a parallel approach based on high priority index feature. In\cite{DCCM}, a one-class WI system is built using directional code co-occurrence matrix feature, along with a dissimilarity measures threshold. As pointed out in \cite{survey}, however, no handcrafted features are particularly suitable for signature verification. On the one hand, the verification methods using hand-crafted features proves to be dependable in cases where a few reference signatures are available. On the other hand, they are also vulnerable to the randomness in the strokes of signatures and have limitation depending on the specific characteristics of the features they extract.

\subsubsection{Learning Features}Recently, the deep neural network based methods have attracted increasing attention. They learn good feature representations directly from signature images, and achieve a large improvement in performance. MCS~\cite{MCS} combines graph edit distance with neural network to build a multi-classifier system. In \cite{BiLSTM} , a combination of CNN and BiLSTM was utilized to tackle the high intra-class variance, and trained by randomly selecting signatures. A thorough study for feature extraction method based on sparse representation were presented in \cite{SR} and achieved promising results . Omid Mersa\cite{transfer} adopted a transfer learning strategy with fine-tuning tactics for off-line signature verification and used the network pretrained in Persian Handwritten datasets, then further transferred it to signature verification. A graph convolutional network based framework is first proposed in \cite{vision} and achieved competitive performance. Signet \cite{cnn2} uses the data of multiple users to train a Convolutional Neural Network (CNN), and then employs the CNN to extract features from the data of individual users. The outputs are then fed to train a WD classifier. Signet achieved the state-of-the-art performance in this category. 

Our proposed method falls into the second category. It employs a new model for feature extraction, and achieves higher performance than the existing methods.

\section{Our Method}
 \begin{figure}[h]
	\centering
	\scalebox{0.18}[0.18]{\includegraphics {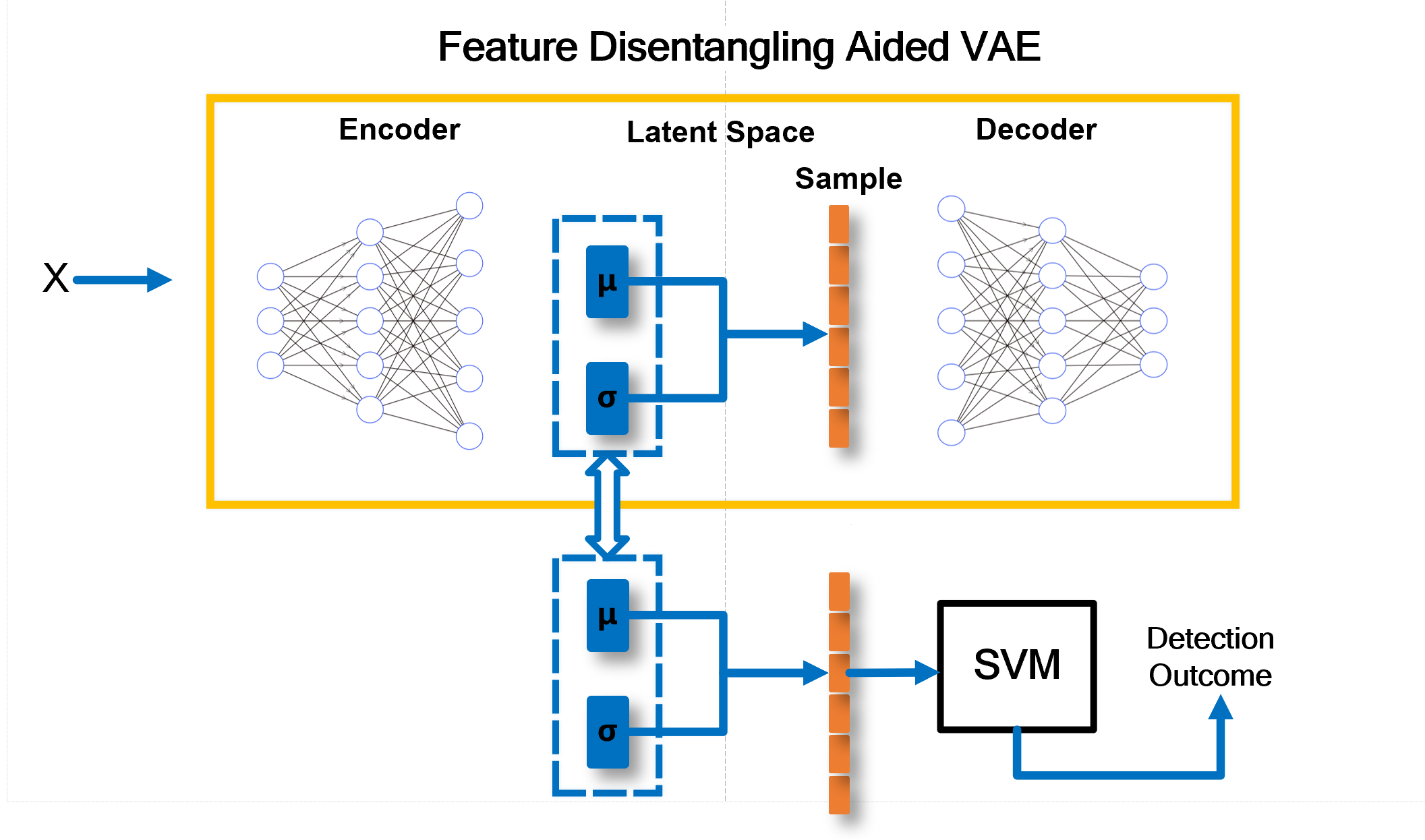}}
	\caption{The architecture of our model.}
\end{figure}

Some of existing signature verification methods are writer-independent (WI), which train a universal classifier for \textit{all} users. However, our method is writer-dependent (WD). That is, it trains a classifier for \textit{every} user. We choose the WD methods due to two reasons. First, they usually have higher performance, although they suffer from higher complexity. Secondly, Writer-Independent (WI) methods are often not suitable for large-scale systems that encompass thousands of users, especially when the users are allowed to dynamically join or leave their system and multiple languages are allowed. 

\subsection{preprocessing}
Some preprocessing steps are necessary in the experiment as neural network expects the inputs to have a fixed size. We first use the OTSU's algorithm\cite{otsu} to automatically find the optimal threshold of pixel intensity between foreground and background so that pixels which has intensity above the threshold will be set to white while pixels with intensity below the threshold remain unchanged in this process. Then we invert the image to make the background to have intensity zero, finally all the inputs are normalized and resized to a fixed size to feed into the neural network.

\subsection{Variational Auto-encoder}
Our proposed model consists of a VAE-based feature extractor and an SVM-based classifier, as shown in Fig. 2.

VAEs are often used to construct complex generative models. A VAE has two components, i.e.,  an encoder and a decoder~\cite{autoencoding}. The encoder encodes its input into a Gaussian distribution, and the decoder attempts to reconstruct the input with a latent vector sampled from the Gaussian distribution. VAEs are good at learning the distribution of its inputs. This motivates us to extract features based on the latent vectors. And these extracted features will be fed to the classifier for signature verification. The structure of a traditional VAE is as below.
\begin{figure}[ht]
	\includegraphics[width = .5\textwidth]{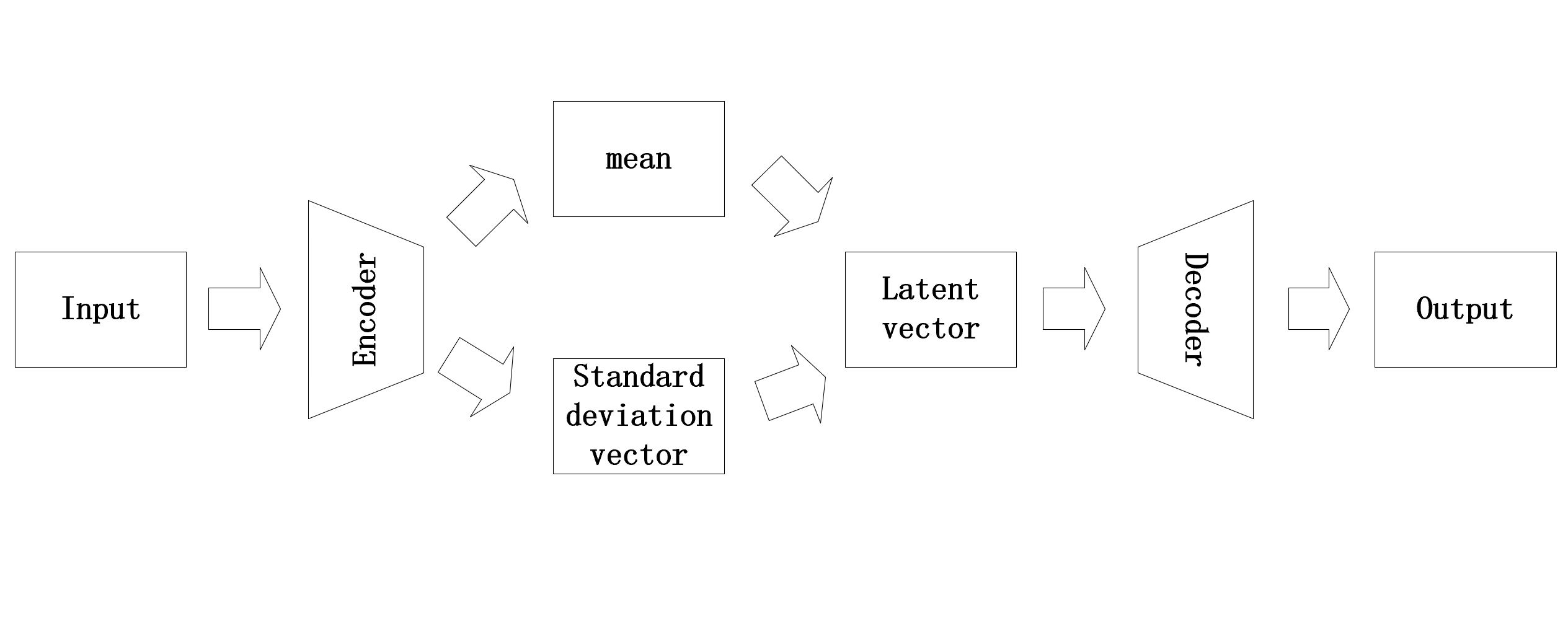}
	\caption{structure for variational auto-encoder}
\end{figure}

	    For each datapoint in a given dataset $\emph{D} = \{{\mathbf x}_{1},...,{\mathbf x}_{N}\}$, a VAE learns to model the underlying distribution of p(z), and the generation process can be portrayed as $ p(\mathbf x)= \int p{(\mathbf x|\mathbf z)}p{\mathbf z}d{\mathbf z}$. However, the computation of marginalization is intractable, thus we represent the marginal log-likelihood as $log{p(x)} = D_{KL}(q_{\phi }({\mathbf z}|{\mathbf x})||p_{\theta }({\mathbf z})) + \mathcal{L}_{VAE}{(\phi,\theta,\mathbf x)}$, where $D_{KL}$	is Kullback-Leibler divergence and  $L_{vae}$ is the variational lower bound of the datapoint.

So we minize the lower bound of the log likelihood \(\mathcal{L}_{VAE}\) via optimizing parameters from encoder and decoder, denoted as $\phi$ and $\theta$ respectively: 
\begin{equation*}{\mathcal{L}_{VAE}}\left( {\phi ,\theta ;{\mathbf x}} \right) = {E_{{q_\phi }\left( {{\mathbf z}|{\mathbf x}} \right)}}\left[ {\log p\left( {{\mathbf x},{\mathbf z}} \right)} \right] - {E_{{q_\phi }\left( {{\mathbf z}|{\mathbf x}} \right)}}\left[ {\log {q_\phi }\left( {{z}|{x}} \right)} \right].\end{equation*}

\subsection{Feature Disentangling for VAE}
Due to their close similarity, however, it may be still hard to distinguish genuine signatures and skilled forgeries in a latent space. To solve this problem, we propose to improve the traditional VAEs by introducing the requirement of feature disentangling. Feature disentangling can enlarge the inter-class distance and shorten the inner-class distance. As will be shown in Section IV, feature disentangling helps separating genuine signatures from forgeries in the latent space, laying the foundation for signature verification at the classifier. 

The requirement of feature disentangling can be met by introducing a new loss functions for VAEs. Accordingly, the VAE in our model have two loss functions, i.e.,  $\mathcal{L}_{VAE}$ and $\mathcal{L}_{FD}$. $\mathcal{L}_{VAE}$ is the loss function of the traditional VAE, which includes reconstruction loss and Kullback-Leibler (KL) divergence. $\mathcal{L}_{FD}$ is designed to disentangle the features of genuine signatures and forgeries, which is given by 
\begin{equation} \label{L_fd}
\mathcal{L}_{FD}= E_{\bm{x}_i,\bm{x}_j\sim  \Omega_{G} \cup \Omega_{F}}f(\bm{x}_i,\bm{x}_j),
\end{equation}
Suppose $d(\bm{x}_i,\bm{x}_j)=(\bm{\mu}_i-\bm{\mu}_j)^2+(\bm{\sigma}_i-\bm{\sigma}_j)^2$. We define $f(\bm{x}_i,\bm{x}_j)=$ as
\begin{equation} \label{new loss}
\left\{
\begin{aligned}
&d(\bm{x}_i,\bm{x}_j) &  y_i=y_j\\
&m-d(\bm{x}_i,\bm{x}_j)&    y_i\neq y_j, d(\bm{x}_i,\bm{x}_j)<m\\
& m& y_i\neq y_j, d(\bm{x}_i,\bm{x}_j)\geq m
\end{aligned}
\right.
\end{equation}
In \eqref{new loss}, $m$ is a positive threshold, $\bm{x}_i$ and $\bm{x}_j$ are the inputs to the model,  and $\Omega_{G}$ and $\Omega_{F}$ are the distribution of genuine signatures and random forgeries, respectively.  $\bm{\mu}_i$ and $\bm{\mu}_j$ are the mean of the Gaussian distribution determined by $\bm{x}_i$ and $\bm{x}_j$, respectively. Similarly,  $\bm{\sigma}_i$ and $\bm{\sigma}_j$ are the variance of the Gaussian distribution determined by $\bm{x}_i$ and $\bm{x}_j$, respectively. $d(\bm{x}_i,\bm{x}_j)=(\bm{\mu}_i-\bm{\mu}_j)^2+(\bm{\sigma}_i-\bm{\sigma}_j)^2$ is introduced to measure the difference between two Gaussian distributions. If the labels for $\bm{x}_i$ and $\bm{x}_j$ are equal (i.e., $y_i=y_j$), the difference $d(\bm{x}_i,\bm{x}_j)$ should be minimized, since the inputs $\bm{x}_i$ and $\bm{x}_j$ belong to the same class. Otherwise, the difference $d(\bm{x}_i,\bm{x}_j)$ should be maximized, since the inputs $\bm{x}_i$ and $\bm{x}_j$ belong to different classes. To prevent $d(\bm{x}_i,\bm{x}_j)$ from being enlarged without limit, we stop enlarging it\footnote{To stop enlarging it, we set $f(\bm{x}_i,\bm{x}_j)$ to a constant, e.g. $m$. In this way, the parameters will not be updated by $\mathcal{L}_{FD}$.} once $d(\bm{x}_i,\bm{x}_j)$ exceeds the threshold $m$. 

After the feature disentangling aided VAE is trained, we can use its encoder for feature extraction. More specifically, we feed it with a signature image $\bm{x}_i$, and get the corresponding features (i.e., feature vector)
\begin{equation} \label{featuer extration}
\bm{f}(\bm{x}_i)=\bm{\mu}_i+\bm{\sigma}_i \cdot \mathcal{N}(\textbf{0},\textbf{I}).
\end{equation}
In \eqref{featuer extration}, $\bm{\mu}_i$ and $\bm{\sigma}_i $ are the outputs of the encoder, and $\mathcal{N}(\textbf{0},\textbf{I})$ represents the standard normal distribution. Here we do not directly use the outputs ($\bm{\mu}_i$, $\bm{\sigma}_i $) as our features. Instead, we introduce a random perturbation by multiplying $\bm{\sigma}_i $ with a sample from $\mathcal{N}(\textbf{0},\textbf{I})$, which makes our classifier more robust.

\subsection{Support Vector Machine}
SVM has been widely used in the case of offline signature verification and its utilization has achieved great performance. In our experiment, we use the Radial Basis Functions(RBF) as kernel function to map non-linear spaces to high-dimensional linearly separable spaces:\begin{equation*} K(x_{i},x_{j})\ =\ \exp(-\gamma\Vert x_{i}-x_{j}\Vert^{2}),\gamma > 0 \end{equation*} 
With RBF kernel function, the resulting SVM classifier function can be rewritten as: 
\[ f({\bf X})=\sum_{k=1}^{N_{i}}y_{k}\alpha_{k}K({\bf X},{\bf S}_{i})+b \eqno{\hbox{(3)}}\] where ${\bf S}_{i}(i=1,2\cdots N)$ are the support vectors from the training samples. Notably, we use the same parameters(\(\gamma\), \(C\)) for all users tested.
 
\subsection{Algorithm}
\begin{algorithm}
	\caption{Feature Disentangling aided VAE based Signature Verification Algorithm (FDV-SV)}
	\textbf{\textit{A. Training}}
	
	\textbf{1. Build the feature extractor}
	\begin{itemize}
		\item  Initialize $\bm{\theta}$, $\eta_1$, $\eta_2$, and $m$;
		\item In each round of training:
		\begin{itemize}
		\item Sample a batch of $(\bm{x}_i, \bm{x}_j)$, denoted by $\mathcal{B}_{GG}$, where $\bm{x}_i \in \Omega_G$ and $\bm{x}_j \in \Omega_G$;
		\item Use $\mathcal{B}_{GG}$ to update $\bm{\theta}$ with gradient descent by
		\begin{equation}
		\bm{\theta} \leftarrow \bm{\theta} - \eta_1\nabla_{\theta}(\mathcal{L}_{VAE})
		 -\eta_2\nabla_{\theta}(\mathcal{L}_{FD})
		\end{equation}		
		\item Sample a batch of $(\bm{x}_i, \bm{x}_j)$, denoted by $\mathcal{B}_{GF}$, where $\bm{x}_i \in \Omega_G$ and $\bm{x}_j \in \Omega_F$;
		\item Use $\mathcal{B}_{GF}$ to update $\bm{\theta}$ with gradient descent by
		\begin{equation}
		\bm{\theta} \leftarrow \bm{\theta} - \eta_1\nabla_{\theta}(\mathcal{L}_{VAE})
		-\eta_2\nabla_{\theta}(\mathcal{L}_{FD})
		\end{equation}		
		\end{itemize}		
	\end{itemize}

	\textbf{2. Train the classifier}
	\begin{itemize}
		\item Feed the encoder with genuine signatures and random forgeries, and use the corresponding outputs from the encoder as positive and negative samples, respectively. 
		\item Train an SVM classifier with the positive and negative samples;
		
	\end{itemize}

	\textbf{\textit{B. Testing}}
	
	\begin{itemize}	
		\item For each $\bm{x}_{i}$ in the testing dataset: 
		\begin{itemize}
			\item Obtain the feature vector $\bm{f}(\bm{x}_i)$ by \eqref{featuer extration}; 
			\item Given $\bm{f}(\bm{x}_i)$, our classifier makes a decision;
		
		\end{itemize}
	\end{itemize}
	
\end{algorithm}

Our algorithm is termed FDV-SV, and its main procedure is described in Algorithm 1, where $\bm{\theta}$ is the parameters of our improved VAE, $\eta_1$ and $\eta_2$ are the learning rates. 


In the training phase, FDV-SV trains a feature disentangling aided VAE for feature extraction, and also trains an SVM-based classifier for signature verification. The training procedure of the feature disentangling aided VAE consists of multiple rounds, each of which has two steps. In step $1$, the algorithm samples a batch of genuine signature pairs, and uses it to update $\bm{\theta}$ according to $\mathcal{L}_{VAE}$ and $\mathcal{L}_{FD}$. In step $2$, it samples a batch of the pairs of genuine signature and random forgery, and uses it to update $\bm{\theta}$ in a similar fashion. Step $1$ helps our model learn the features of genuine signatures, since its inputs are all genuine signatures. Step $2$ guides the model to distinguish genuine signatures from forgeries, since $\bm{x}_i$ and $\bm{x}_j$ ($\bm{x}_j$ is a random forgery) now belong to different classes. 

In the testing phase, FDV-SV first extracts the features from a signature image using the improved VAE model, and then feeds the features to the classifier for signature verification. 

\section{Performance Evaluation}
In this section, we conduct extensive experiments to evaluate our method and compare it with $11$ signature verification methods. 
\subsection{Datasets \& Settings}
Our experiments were conducted over two public and mainstream datasets: MCYT-75 \cite{mcyt} and  GPDS-synthetic \cite{gpds}. MCYT-75 contains the data of $75$ users. For each user, MCYT-75 stores $15$ genuine signatures and $15$ skilled forgeries. GPDS-synthetic has the data of $4000$ users. For each user, GPDS-synthetic contains $24$ genuine signatures and $30$ skilled forgeries. 

\begin{table}[htbp]
	\scriptsize
	\centering
	\caption{Training data and Testing data.}
	\begin{tabular}{c|c|c}
		\toprule
		Datasets&Training data &Testing data \\
		\midrule
		\multirow{2}{*}{MCYT-75}
		& \multicolumn{2}{|c}{all of $75$ users}    \\
		\cline{2-3}
	&$10$ genuine per user&$5$ genuine and $15$ forgeries per user    \\
	\hline
	\multirow{2}{*}{GPDS-synthetic }
			&\multicolumn{2}{|c} {the first $2000$ users}\\
			\cline{2-3}
	      &$12$ genuine per user&$12$ genuine and $30$ forgeries per user    \\
		\bottomrule
	\end{tabular}
\end{table}

We further split the data into two parts: training data and testing data, as described in Table I. For the purpose of fair comparison, we choose $10$ or $12$ genuine signatures as training data for each user in MCYT-75 or GPDS-synthetic, which coincides with the settings in many literature~\cite{cnn2}\cite{MCS}\cite{GED}\cite{LBP}.  In addition, we use the first genuine signatures of each user (except for the user considered in current experiment) in MCYT-75 and the last $1000$ users in GPDS-synthetic as the random forgeries. That is, we have $74$ random forgeries for each user in MCYT-75, and $1000$ random forgeries for each user in GPDS-synthetic.

Furthermore, either the encoder or the decoder used in our experiments has three hidden layers, each of which contains $200$ neurons. Both $\bm{\mu}$ and $\bm{\sigma}$ are $2$-dimension in the experiments discussed in subsection B, and $400$-dimension in the experiments discussed in subsection C. 

\subsection{Feature Disentangling}
We first show that feature disentangling is able to clearly separate genuine signatures from forgeries in a latent space. For the purpose of visualization,  we consider a 2D latent space (i.e., both $\bm{\mu}$ and $\bm{\sigma}$ are 2-dimension). Using the VAE and \eqref{featuer extration}, we obtain the feature vectors (i.e., the latent representations) for some genuine signatures, skilled forgeries and random forgeries, which are shown in Fig. 3.  It can be seen that without feature disentangling, the latent representations for these three kinds of samples are mixed together. When feature disentangling is used (i.e., loss function \eqref{L_fd} is introduced), these representations become clearly separated, as depicted in subfigure 3(b). These results confirm that feature disentangling helps our model achieve more discriminative features. 
\begin{figure}[h]
	\centering
	\subfigure[]{
		\begin{minipage}{4cm}
			\centering
			\includegraphics[scale=0.28]{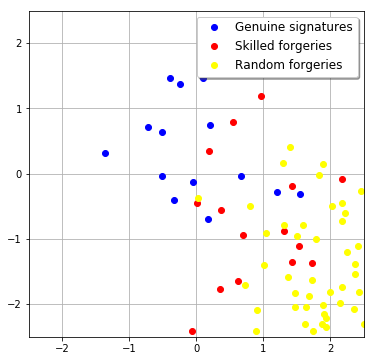}
		\end{minipage}%
	}%
	\subfigure[]{
		\begin{minipage}{4cm}
			\centering
			\includegraphics[scale=0.28]{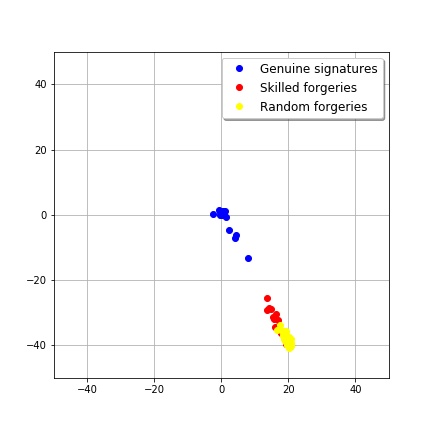}
		\end{minipage}%
	}%
	\centering
	\caption{The feature vectors of three kinds of samples. In (a), feature disentangling is absent. In (b), feature disentangling is used.}
\end{figure}

\subsection{Detection Performance}
Now, we compare our method with $11$ representative signature verification methods. To this end, we use three classical metrics, i.e., False Rejection Rate (FRR), False Acceptance Rate(FAR) and Equal Error Rate (EER). FRR is the ratio of genuine signatures misclassified as forgeries, and FAR represents the ratio of the forgeries misclassified as genuine signatures. EER corresponds to the common value when FAR and FRR are equal. The lower the EER value is, the higher accuracy a signature verification system achieves.

The experimental results of our method on two datasets are given in Tables II and III, respectively. For comparison, we also provide the results of the other $11$ methods in these two tables. The second column of both tables represents the number of genuine signatures used by model training. The notation ``-" indicates that the corresponding result is not provided by the authors. Tables II and III demonstrate that our method significantly outperforms all the competitors. Among them, Signet \cite{cnn2} achieved the best result (i.e., EER=$2.87\%$) on MCYT-75. Our method succeeds in bringing down this record by $53\%$, and obtains an ultra low EER value of $1.35\%$. In addition, the multi-task contrastive model proposed in \cite{multi} obtained the lowest EER value $4.52\%$ on GPDS-synthetic. Our method dramatically reduces this value to $2.22\%$. 

\begin{table}[htbp]
	\scriptsize
	
	\centering
	\caption{Performance comparison on MCYT-75}
	
	\begin{tabular}{cccccc}
		\toprule
		Methods&Ref&FRR&FAR&EER \\
		\midrule
		
		BoVM-VLAD-KAZE \cite{BVK}&10&-   &-           &6.40\%    \\
		MCS \cite{MCS}     &10   &4.00\%   &17.24\%     &9.16\%    \\
		Signet \cite{cnn2}  &10   &-      &-            &2.87\%    \\
		DMML \cite{DMML}    &10   &10.73\%  &10.02\%      &10.06\%   \\
		MT-SigNet (NT-Xent) \cite{multi}  &10   &-\%  &-\%      &3.95 \%   \\
		Transfer learning \cite{transfer} &10 &- &- &3.98\%           \\
		Archetypes \cite{archetype}&5  &3.97\%   &3.97\%        &3.97\%    \\    
		Genetic \cite{ga} &10 &6.25\%  &5.67\%  &- \\
		Scored-level fusion \cite{LBP}   &10   &8.59\%   &6.77\%        &7.08\%    \\

		\textbf{Ours}&\textbf{10}  &\textbf{0.42\%}   &\textbf{1.64\%}      &\textbf{1.35\%}    \\
		
		\bottomrule
	\end{tabular}
\end{table}

\begin{table}[htbp]
	\scriptsize
	\centering
	\caption{Performance comparison on GPDS-synthetic}
	
	\begin{tabular}{cccccc}
		\toprule
		Methods&Ref&FRR&FAR&EER \\
		\midrule
		GED \cite{GED} &10           &- &- &8.29\%  \\
		Cylindrical Shape \cite{CS} &12 &3.51\%      &13.91\%          &-       \\
		Transfer Learning \cite{transfer}&10&-   &-           &6.81\%    \\
		
		Boundary pixels \cite{pixels}   &12   &12.40\%  &14.84\%      &12.57   \\
		CNN-BiLSTM \cite{BiLSTM}     &10   &-   &-    &16.44\%   \\
		
		Shape Correspondence \cite{SC}&12   &10.65\%      &10.19\%        &-       \\
		MT-SigNet (NT-Xent) \cite{multi}   &12   &-\%      &-\%        & 4.52       \\
		
		Genetic \cite{ga} &12 &4.16\%  &5.42\%   &- \\
		MCS \cite{MCS}  &10   &5.05\%      &9.11\%            &7.29\%    \\
		
		\textbf{Ours}&\textbf{12}  &\textbf{0.02\%}   &\textbf{3.75\%}     &\textbf{2.22\%}    \\
		
		\bottomrule
	\end{tabular}
\end{table}

\section{Conclusion}
In this paper we propose the offline signature verification method based on variational auto-encoder. The key point of this method is that feature disentangling is introduced to guide a VAE to extract discriminative features directly from signature images. Extensive experiments show that our method significantly surpasses other representative signature verification methods, and achieves a great improvement in state-of-the-art performance. 

In our future work, we will focus on exploring unsupervised learning based on Variational Autoencoders (VAE) to further reduce complexity and enhance accuracy.Additionally, we intend to explore the potential application of Variational Autoencoders (VAE) for signature generation in the context of data augmentation, with the goal of enhancing the accuracy and effectiveness of the verification process.
\bibliographystyle{ieeetr} 
\bibliography{ywcite_new}

\end{document}